\documentclass[lettersize,journal]{IEEEtran}
\usepackage{amsmath,amsfonts}
\usepackage{algorithmic}
\usepackage{algorithm}
\usepackage{array}
\usepackage[caption=false,font=small,labelfont=sf,textfont=sf]{subfig}
\usepackage{textcomp}
\usepackage{stfloats}
\usepackage{url}
\usepackage{verbatim}
\usepackage{graphicx}
\usepackage{cite}
\usepackage{multirow}
\usepackage{multicol}
\usepackage{xcolor}

\usepackage{todonotes}

\begin{document}

\title{Safe Semi-Supervised Contrastive Learning Using In-Distribution Data as Positive Examples}

\author{Min~Gu~Kwak, Hyungu~Kahng,~and~Seoung~Bum~Kim{}
\thanks{M. G. Kwak is with the School of Industrial and Systems Engineering, Georgia Institute of Technology, Atlanta, Georgia, USA (e-mail: mkwak35@gatech.edu)}
\thanks{H. Kahng is with the Department of Digital Management, Korea University Sejong Campus, Sejong, Republic of Korea (e-mail: hgkahng@korea.ac.kr)}
\thanks{S. B. Kim is with the School of Industrial Management Engineering, Korea University, Seoul, Republic of Korea (e-mail: sbkim1@korea.ac.kr)}}

\markboth{DRAFT}%
{Shell \MakeLowercase{\textit{et al.}}: A Sample Article Using IEEEtran.cls for IEEE Journals}


\maketitle

\begin{abstract}
Semi-supervised learning methods have shown promising results in solving many practical problems when only a few labels are available. The existing methods assume that the class distributions of labeled and unlabeled data are equal; however, their performances are significantly degraded in class distribution mismatch scenarios where out-of-distribution (OOD) data exist in the unlabeled data. Previous safe semi-supervised learning studies have addressed this problem by making OOD data less likely to affect training based on labeled data. However, even if the studies effectively filter out the unnecessary OOD data, they can lose the basic information that all data share regardless of class. To this end, we propose to apply a self-supervised contrastive learning approach to fully exploit a large amount of unlabeled data. We also propose a contrastive loss function with coefficient schedule to aggregate as an anchor the labeled negative examples of the same class into positive examples. To evaluate the performance of the proposed method, we conduct experiments on image classification datasets—CIFAR-10, CIFAR-100, Tiny ImageNet, and CIFAR-100+Tiny ImageNet—under various mismatch ratios. The results show that self-supervised contrastive learning significantly improves classification accuracy. Moreover, aggregating the in-distribution examples produces better representation and consequently further improves classification accuracy.
\end{abstract}


\section{Introduction}
\label{sec:introduction}
Deep neural networks have shown promising results in various supervised learning problems, including image classification \cite{he2016deep}, object detection \cite{gu2018recent}, natural language processing \cite{li2020survey}, and signal data analysis \cite{zhang2021unsupervised}. It is well known that a large-scale training dataset with well-annotated labels is required for dependable performance in supervised learning tasks \cite{lecun2015deep}. However, creating such extensive collections of labeled data is typically time-consuming and incurs high costs for many real-world problems. Consequently, it limits the broad adoption of deep neural networks for many practical issues and applications.

Semi-supervised learning (SSL) algorithms have been proposed to reduce the labeling overload and improve model performance by leveraging unlabeled data when only a limited number of data samples have corresponding labels \cite{chapelle2009semi, nie2019multiview}. SSL algorithms generally learn the manifold data representation from a large amount of unlabeled data and classify new observations into classes with suggestions from a small amount of labeled data. The adoption of the SSL framework for deep neural networks has shown great results in many real-world applications. Deep SSL advances include but are not limited to transportation \cite{dabiri2019semi}, spam detection \cite{qiu2020adaptive}, medical data analysis \cite{ito2019semi}, air quality forecasting \cite{han2022semi}, and speech recognition \cite{ling2020deep}. The methodological developments of SSL can be divided into several categories, such as entropy minimization \cite{grandvalet2004semi}, consistency regularization \cite{sajjadi2016regularization, laine2016temporal, tarvainen2017mean, miyato2018virtual}, and holistic methods with strong data augmentation techniques \cite{berthelot2019mixmatch, berthelot2019remixmatch, sohn2020fixmatch}.

\begin{figure}[t]
\centering
    \subfloat[Conventional SSL]{\includegraphics[width=1.00\linewidth, keepaspectratio]{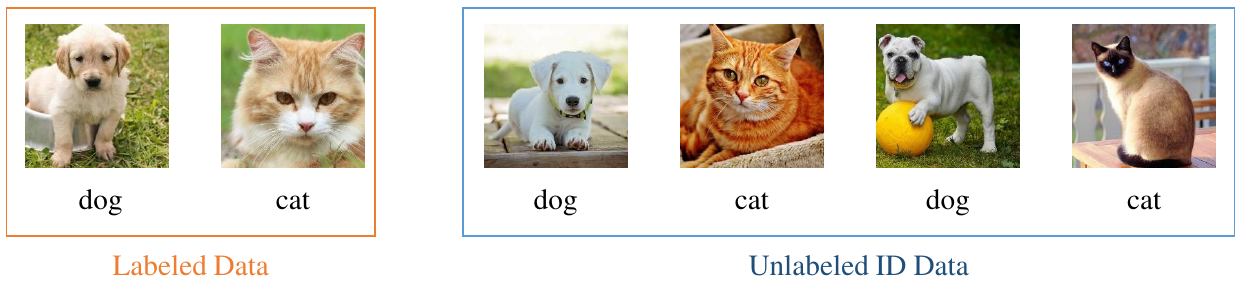}\label{fig:mismatch_a}}
    \\[1.1ex]
    \subfloat[Class distribution mismatch in SSL]{\includegraphics[width=1.00\linewidth, keepaspectratio]{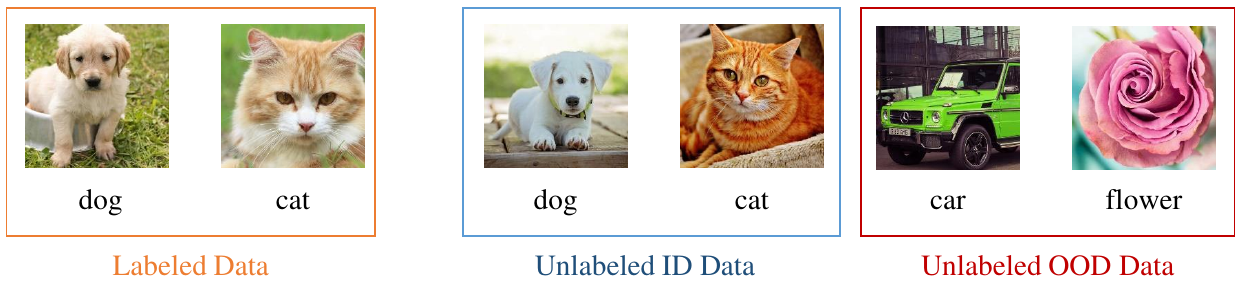}\label{fig:mismatch_b}}
    \caption{An example of when the identical class assumption is violated in SSL. OOD data (car and flower indicated with red box) exists in unlabeled data that are not present in the labeled data.}
    \label{fig:mismatch}
\end{figure}

One of the common and critical assumptions of SSL is that the class distributions of labeled and unlabeled data are identical. In other words, out-of-distribution (OOD) data, which are different from target classes or in-distribution (ID) data, should not exist in unlabeled data. However, this assumption often fails to hold in real-world scenarios \cite{yang2011can}. In practice, when gathering vast quantities of data, it is nearly impossible to thoroughly inspect each sample. Consequently, the inclusion of OOD data becomes an inevitable part of the dataset. This is called the class distribution mismatch problem. Note that the data having the same distribution as the labeled data is called \textit{target} or ID data, and data having a different distribution is called \textit{mismatch} or OOD data. Figure \ref{fig:mismatch} illustrates an example of the mismatch problem in an image classification task. In an SSL problem of classifying dogs and cats, OOD data such as cars and flowers may exist in the unlabeled data. The performances of deep SSL methods have been shown to be significantly hindered by the existence of unlabeled OOD data. Moreover, a high mismatch ratio makes SSL algorithms perform inferiorly, even to a model trained only with a small amount of labeled data \cite{oliver2018realistic, chen2020semi, guo2020safe}. Only recently has the attention on addressing the practical and realistic scenario been highlighted, and several studies have been conducted.

\begin{figure}[t]
    \centering
    \includegraphics[width=0.90\linewidth, keepaspectratio]{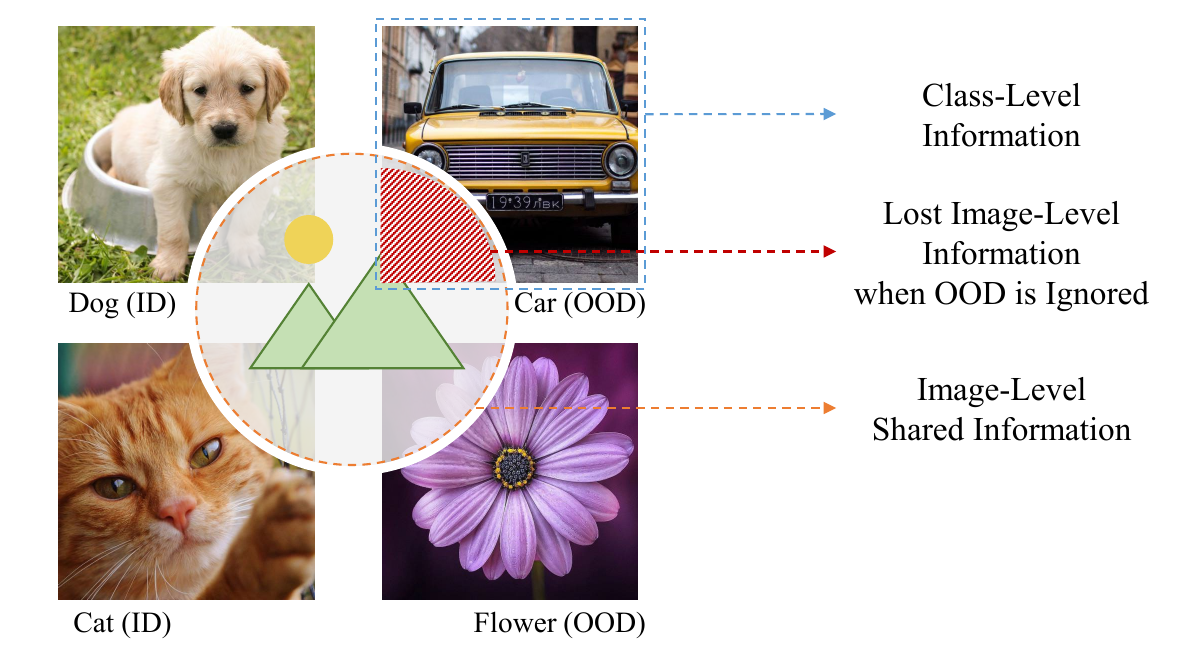}
    \caption{Simplified semantic data structure of natural images. Two key features exist: (1) class-level information (e.g., a dog's nose, a cat's ears) and (2) image-level information (e.g., backgrounds, textures). The red area indicates potential loss of class-level information when OOD data is excluded.}
    \label{fig:data_structure}
\end{figure}

Safe SSL is a research area that addresses the class distribution mismatch problem in SSL. Its primary focus is on leveraging ID data while making the OOD data less likely to affect training. Typical approaches in safe SSL involve either removing OOD data from the training set \cite{chen2020semi} or imposing proper weight on each unlabeled sample \cite{guo2020safe}. However, it is important to be aware that using additional methods to remove OOD data might also accidentally erase valuable ID data. Losing ID data could harm the accurate classification of target classes. Moreover, we argue that ignoring OOD data in training might lose valuable basic information that all collected data share regardless of class. For example, the semantic information of natural images can be simplified, as demonstrated in Figure \ref{fig:data_structure}. Each image possesses distinct class-level information crucial for classification \cite{zhou2016learning}, such as a dog's nose or a cat's ears. Concurrently, these images also have image-level information that represents the general semantic data structure of a natural image. This includes universal features like natural backgrounds, lighting and shadow patterns, and a spectrum of natural colors and textures. These common elements are important for deep learning models, aiding in the comprehensive understanding and differentiation of images for tasks such as image classification and object detection. Even if safe SSL methods filter out the unnecessary OOD class information, they can lose some portion of the shared information, as highlighted in red in Figure \ref{fig:data_structure}.

Self-supervised contrastive learning (SSCL) has recently made significant advancements in uncovering complex intrinsic patterns in unlabeled data \cite{liu2021self}. Recent promising methods have made significant improvements in many tasks, such as image classification \cite{caron2020unsupervised}, anomaly detection \cite{zhang2022adaptive, zheng2021generative}, graph data analysis \cite{wu2021self}, and so on. At its core, SSCL is built upon the principles of instance discrimination and invariant mapping \cite{wu2018unsupervised, hadsell2006dimensionality}, which involve distinguishing each instance from others while maintaining consistency across data augmentations. Specifically, SSCL utilizes the relationships between anchors, positive examples, and negative examples for instance discrimination. Here, an anchor refers to the target instance being distinguished, a positive example is a different augmented view originating from the same instance as the anchor, and a negative example is a view originating from a different instance. The latest studies have successfully captured the semantic structure hidden in the data and produced general representations that can be broadly used for various downstream tasks. Unlike SSL methods that use both labeled and unlabeled data in a single training phase, the SSCL methods are trained without labels in the pre-training phase and then proceed to downstream tasks in the fine-tuning phase with only labeled data. In particular, the methods do not need to consider the class distribution mismatches.

In this study, we propose a safe semi-supervised contrastive learning method to utilize all the unlabeled data in a class distribution mismatch scenario. We hypothesize that the SSCL method, built upon the concept of instance discrimination, can provide proper initial network parameters for classification while avoiding the negative effect of mismatching. Also, the proposed method does not filter the unlabeled OOD data, so it can learn the general data representation well without information loss. Moreover, we introduce a loss function and its coefficient schedule, which generates more appropriate representations for safe SSL. The main idea of the proposed loss function involves specifically reassigning those labeled negative examples that share the same class label as the anchor, treating them as additional positive examples. The selectively reassigned examples are certainly ID. It aims to improve the model's capability to recognize and aggregate similar patterns within a class, thus enhancing the accuracy and distinctiveness of class representations. This satisfactorily clusters the ID classes during the pre-training phase. The coefficient schedule gradually reduces the influence of the proposed loss function on model training to avoid overfitting problems that may occur by reusing a small amount of the labeled data used in the pre-training phase for the fine-tuning phase. We demonstrate the effectiveness of our method with experiments on benchmark datasets—CIFAR-10, CIFAR-100, Tiny ImageNet, and CIFAR-100+Tiny ImageNet—under various mismatch ratios. To the best of our knowledge, the present study is the first attempt to applying a SSCL approach to safe SSL without strong data augmentations in pre-training.

\section{Related Works}
\label{sec:related works}
Traditional deep SSL methods have raised the model capability to the point where it can achieve similar performance to supervised learning. They tackle the scenario when the class categories of labeled data and unlabeled data are identical. Pseudo-label was proposed as a method to use predicted class probabilities as targets for a standard supervised loss function \cite{lee2013pseudo}. Making the network produce confident predictions by minimizing the entropy for all unlabeled data has also been proposed \cite{grandvalet2004semi}. The $\Pi$-Model introduced a simple framework to apply consistency regularization in SSL \cite{sajjadi2016regularization}. Consistency regularization means that even when subtle perturbations such as data augmentation are given to an input, a model should produce an output similar to that of the original input. Temporal ensembling was proposed as the second version of $\Pi$-Model, which aggregates past predictions to derive current predictions for improving classification accuracy and training speed \cite{laine2016temporal}. An additional momentum-updated teacher network was introduced by mean teacher to enable the SSL model to produce stable predictions \cite{tarvainen2017mean}. Virtual adversarial training (VAT) was proposed to enhance the model's robustness against local perturbations, which are approximated to those that most significantly affect the predictions during training \cite{miyato2018virtual}. Another study found that stochastic weight averaging (SWA) improved generalization performance by averaging network parameters along the trajectory of stochastic gradient descent \cite{athiwaratkun2018there}. Because the experiments to test these methods were conducted in different experimental settings, a unified evaluation protocol for SSL was provided \cite{oliver2018realistic}. Most recently, holistic approaches of UDA \cite{xie2020unsupervised}, MixMatch \cite{berthelot2019mixmatch}, and FixMatch \cite{sohn2020fixmatch} have yielded performance improvements inspired by strong data augmentations. However, the existing SSL methods are built on a strict assumption that there is no OOD data in the unlabeled data. This assumption is readily violated in many realistic circumstances, resulting in significant performance degradation.

Several safe SSL methods have been proposed to tackle the class distribution mismatch problem by focusing on utilizing ID data to eliminate or reduce the negative impact of OOD data. The uncertainty-aware self-distillation (UASD) proposed a self-distillation method to dynamically filter out OOD data based on confidence scores calculated during training \cite{chen2020semi}. It stabilized the filtering process by averaging the confidence scores of past predictions on the validation dataset. DS3L adopted a bi-level meta learning optimization framework to impose the proper weight on the unlabeled data to improve generalization performance \cite{guo2020safe}. It used a single-layer network updated by only labeled ID data to decrease the influence of unlabeled OOD data. Multi-task curriculum framework (MTCF) was a proposed end-to-end trainable framework that detects OOD data and classifies ID data simultaneously \cite{yu2020multi}. OpenMatch combined the learning process of FixMatch and novelty detection to learn ID data representations while rejecting OOD data \cite{saito2021openmatch}. Semi-supervised open set classification (S2OSC) proposed a framework of simultaneously filtering the distinct OOD data and assigning new super-class labels to them. S2OSC also adopted knowledge distillation to effectively separate the remaining uncertain unlabeled data \cite{yang2021s2osc}. The performances of the abovementioned methods might lose the basic representations that OOD data possess, regardless of class. Contrary to the studies that consider OOD data to be totally harmful, transferable OOD data recycling (TOOR) proposed to reuse some OOD data having informative information by employing adversarial domain adaptation method \cite{huang2022they}. The performances of the existing methods necessarily depend on the OOD data's detectability. We demonstrate that our method can improve classification performance by adopting SSCL approach without filtering the unlabeled OOD data.

Recently, contrastive learning has gained great attention in self-supervised learning because of its ability to learn generalized representations without labels \cite{jaiswal2021survey}. The main idea of SSCL is to make the representations from different data augmentations on an instance similar to but distinct from those of other instances. Noise contrastive estimation was successfully applied to compare instances rather than classifying them into pre-defined classes \cite{wu2018unsupervised}. For high-dimensional time series data, an autoregressive model with a contrastive loss that predicts the future latent space was proposed \cite{oord2018representation}. Pretext-invariant representation learning (PIRL) uses a memory bank to store representations of all instances that are used for efficient negative sampling during training \cite{misra2020self}. To address the memory inefficiency caused by large batch sizes or a memory bank with a memory queue, momentum contrast (MoCo) was proposed \cite{he2020momentum, chen2020improved}. This method also applies a momentum or exponential moving average update for the network to achieve consistency in the representation during training. SimCLR uses a simple end-to-end architecture that performs well through simultaneous negative sampling from large batch sizes \cite{chen2020simple, chen2020big}. Several most recent studies successfully leveraged strong data augmentations. For example, SwAV used a multi-crop strategy to produce more views of an instance to increase the number of data augmentations along with a novel cluster-based contrastive loss function \cite{caron2020unsupervised}. DINO also used the multi-crop strategy in synergy with vision transformer (ViT) \cite{caron2021emerging}. Although the existing methods achieved much success, they have not yet been applied to the class distribution mismatch problem in SSL.

\section{Proposed Method}
\label{sec:proposed method}
We propose a safe semi-supervised contrastive learning method to mitigate the performance degradation caused by class distribution mismatch. The core principle of our method is the specific reassignment of labeled negative examples that share the same class label as the anchor, treating them as additional positive examples. The reassigned examples are indeed ID. Our method enhances the model's ability to identify and aggregate similar patterns within a class, thereby increasing the precision and distinctiveness of class representations. Additionally, by not discarding OOD data and instead utilizing it in our training process, we can more effectively learn the information commonly shared across the dataset.

The proposed method is based on MoCo, an SSCL approach renowned for its enhanced representation learning capability and computational efficiency. It is worth noting that the selection of MoCo as our foundation is attributed to its dynamic dictionary mechanism of a memory queue. MoCo maintains a memory queue of encoded representations, which are employed as negative examples. In the context of SSL, where labeled data is scarce, methods like SimCLR that do not utilize a memory queue require significantly larger batch sizes. This is because, in smaller batches, the number of negative examples with labels can be extremely low or even non-existent. However, large batch sizes often lead to memory shortage issues. To effectively implement our method, we designed our model based on the MoCo framework, which reduces the need for large batch sizes while ensuring a sufficient supply of labeled negative examples for contrastive learning in class distribution mismatch scenarios.

\begin{figure*}[t]
    \centering
    \includegraphics[width=0.90\linewidth, keepaspectratio]{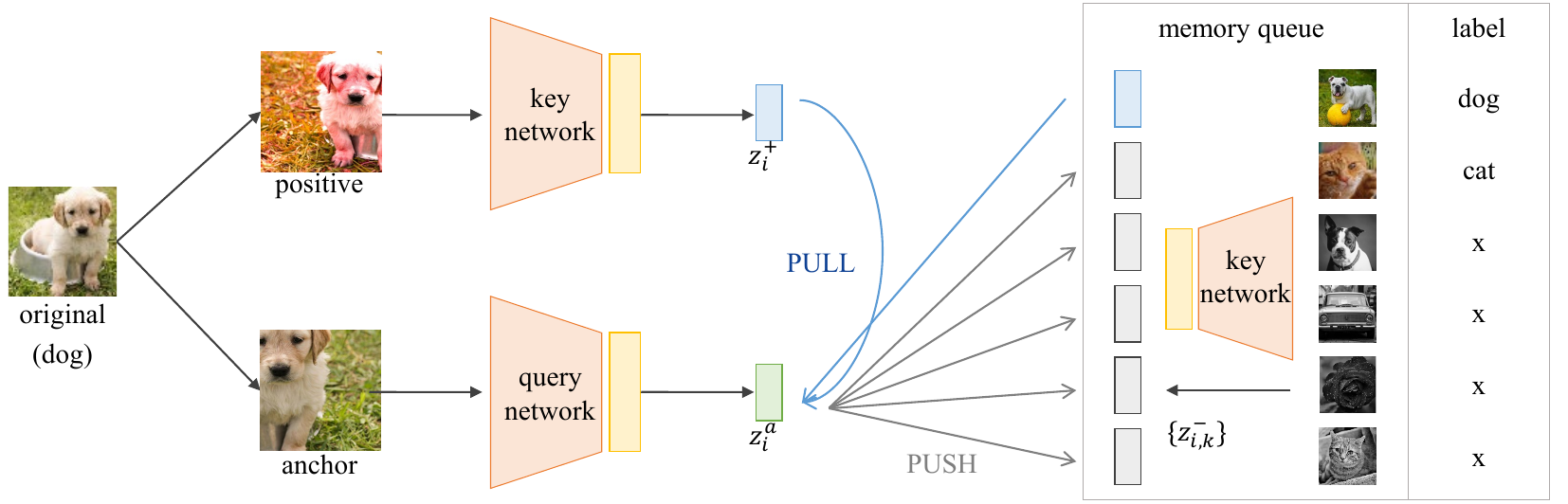}
    \caption{Graphical overview of the proposed method that leverages the labeled data among negative examples in a memory queue as positive examples.}
    \label{fig:proposed_method}
\end{figure*}

In this section, we describe a problem statement, the instance discrimination process of SSCL based on MoCo, and the proposed contrastive loss function and its coefficient schedule suitable for SSL in the class distribution mismatch scenario. Figure \ref{fig:proposed_method} shows an overview of the proposed method.

\subsection{Problem Statement}
\label{sec:problem statement}
Let $\mathcal{D}=\mathcal{D}_L \cup \mathcal{D}_U$ be a training dataset, where $\mathcal{D}_L=\{(x_i, y_i)\}_{i=1}^{n}$ is the labeled data of $n$ instances and $\mathcal{D}_U=\{x_j\}_{j=1}^{m}$ is the unlabeled data of $m$ instances. In a typical SSL classification problem, $x \in X \in \mathbb{R}^d$, $y \in \{1, 2, \cdots, C\}$, and $m \gg n$ where $d$ is the number of input dimension and $C$ is the number of classes. The goal of deep SSL is to train a parameterized network $h$ that minimizes the following loss function:
\begin{equation}
\label{eqn:ssl}
    \mathcal{L}_\textit{SSL}=
    \sum_{{(x_i, y_i) \in \mathcal{D}_L}}l_1(h(x_i, y_i))+
    \sum_{{x_j} \in \mathcal{D}}l_2(h(x_j)),
\end{equation}
where $l_1$ and $l_2$ are loss functions. In general, cross-entropy is used for $l_1$, and a regularization function is used for $l_2$. The network $h$ converges to capture the data representations from $D$ and classifies new instances with a clue from the limited number of $\mathcal{D}_L$. However, if the class distribution of $\mathcal{D}_L$ and $\mathcal{D}_U$ is different, the data representations are poorly learned, resulting in the classification performance degradation \cite{oliver2018realistic}.

\subsection{SSCL: Instance Discrimination}
\label{sec:instance discrimination}
SSCL methods conduct instance discrimination, enabling a model to utilize all information in unlabeled data without being adversely influenced by the distribution discrepancies between labeled and unlabeled data. The instance discrimination aims to learn the representations by making similar instances pull each other together and dissimilar instances push each other apart. To achieve this goal, stochastic data augmentation strategies are applied. Given an instance $x_i \in \mathcal{D}$, two different views are generated by applying data augmentation twice: $x_i^a$ and $x_i^+$ denote the anchor example and positive example, respectively. Then, data augmentation is applied to $K$ instances sampled from $\mathcal{D}-\{x_i\}$ to achieve negative examples $\{x_{i,k}^-\}_{k=1}^K$. The representations of a positive pair $(x_i^a, x_i^+)$ must be close because they originate from an identical instance. On the contrary, the representations of any negative pair $(x_i^a, x_{i,k}^-)$ must be far away. Through a neural network $f$ that embeds an input $x$ to a representation vector $z=f(x)\in\mathbb{R}^{d'}$, representations of positive and negative pairs can be achieved, denoted by $(z_i^a,z_i^+)$ and $(z_i^a, z_{i,k}^-)$, respectively. The output vector $z$ is normalized by its L2-norm \cite{wu2018unsupervised}. Consequently, the contrastive loss function of instance discrimination is formulated as follows:
\begin{equation}
\label{eqn:moco_loss}
    \mathcal{L}_{i}^{MoCo} = -\log\tfrac
    {\exp({{z_i^a}\cdot{z_i^+}}/\tau)}
    {
        \exp({{z_i^a}\cdot{z_i^+}}/\tau) + \sum_{k=1}^{K}\exp({z_i^a}\cdot{z_{i,k}^-}/\tau)
    },
\end{equation}
where $\tau$ is a temperature hyperparameter for scaling. Because the representation vector $z$ is L2-normalized, the inner dot product can easily measure the similarity of a pair's corresponding components. From the instance discrimination point of view, minimizing $\mathcal{L}_{i}^{MoCo}$ is equivalent to maximizing the probability of assigning an anchor example to a positive example among $K+1$ instances consisting of one positive example and $K$ negative examples.

To compare the anchor with other examples, following the MoCo framework, we utilized two networks: key network $f_{\theta_k}$ and query network $f_{\theta_q}$, where $\theta_k$ and $\theta_q$ denote the corresponding weight parameters. The networks are trained to find the similar representation of query (anchor) from keys (positive and negative examples); namely, $z_i^a=f_{\theta_q}(x_i^a)$, $z_i^+=f_{\theta_k}(x_i^+)$, and $z_{i,k}^-=f_{\theta_k}(x_{i,k}^-)$ can be obtained. Specifically, $\theta_k$ is not backpropagated after every iteration. Instead, it is updated by an exponential moving average or momentum update of $\theta_q$. The update rule is formulated as follows:
\begin{equation}
\label{eqn:momentum}
    \theta_k \leftarrow m\cdot\theta_k + (1-m)\cdot\theta_q,
\end{equation}
where $m\in[0,1)$ is the momentum coefficient. $m$ is usually set by a large value, such as 0.950 or 0.999. The slightly changing key network preserves the consistency of key representations. It makes the instance discrimination task difficult enough, and consequently, the model converges to generate proper representations \cite{he2020momentum}.

\begin{figure}[t]
\centering
    \subfloat[Unit hypersphere]{\includegraphics[width=0.60\linewidth, keepaspectratio]{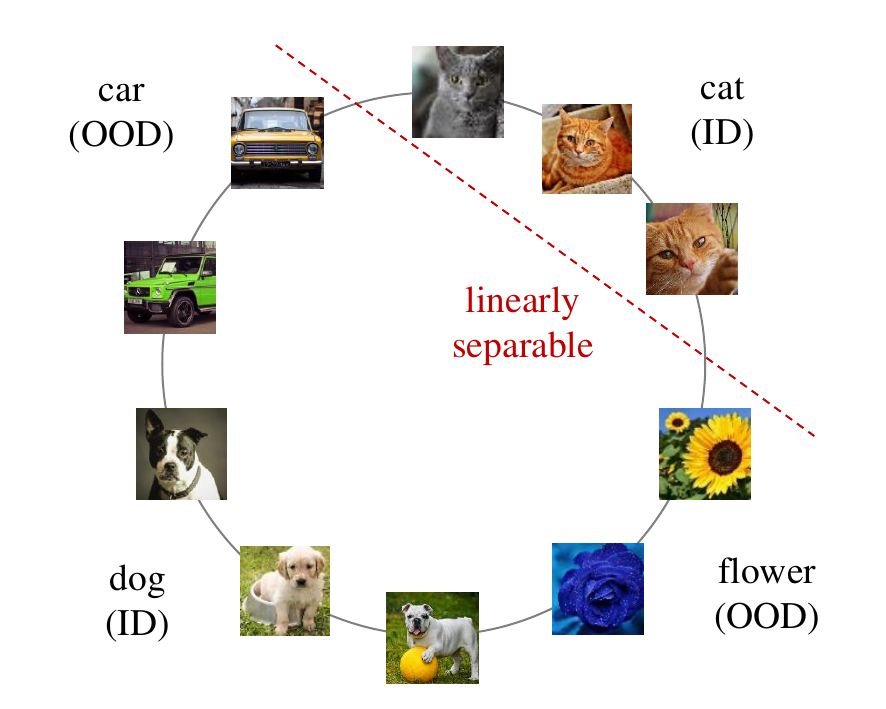}\label{fig:3a}}
    
    \subfloat[Desired hypersphere for safe SSL]{\includegraphics[width=0.60\linewidth, keepaspectratio]{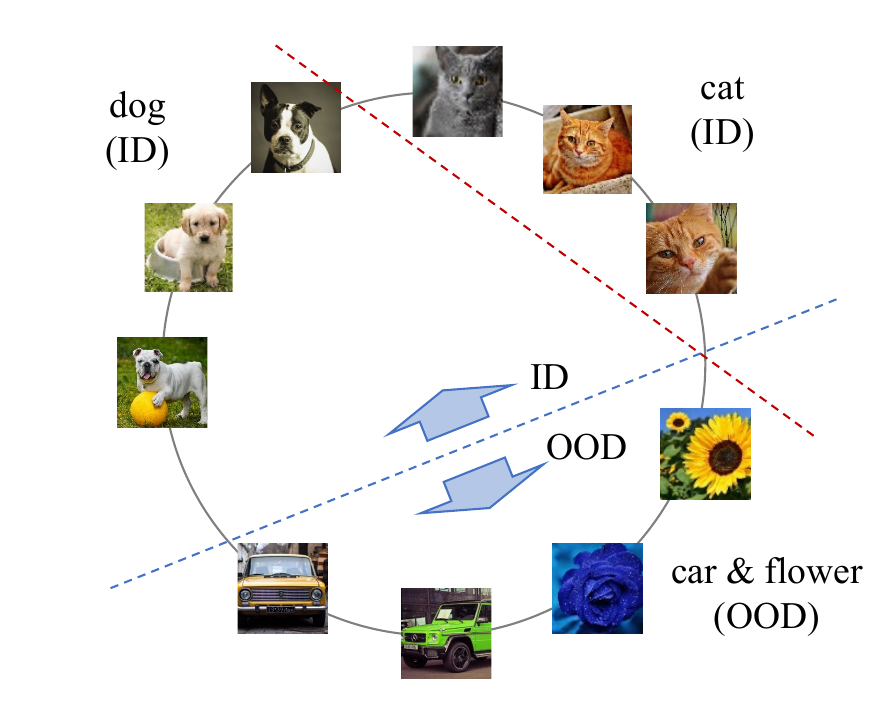}\label{fig:3b}}
    \caption{Learned representations on an L2-normalized unit hypersphere. (a) A hypersphere that classes are linearly separable; (b) A hypersphere with additional characteristics to keep ID and OOD apart.}
    \label{fig:3}
\end{figure}

\subsection{Proposed Contrastive Loss and Schedule Method}
\label{sec:proposed loss}
It is known that the classes are linearly separable on an L2-normalized unit hypersphere if the instances are successfully discriminated and semantically well-clustered (Figure \ref{fig:3a}). Therefore, if desired, we can simply use MoCo to provide initial parameters for SSL. However, we expect the model to have an additional characteristic to separate ID data from OOD data because we consider the SSL under the class distribution mismatch (Figure \ref{fig:3b}). The model should provide representations suitable for performing the downstream classification problem, and the OOD data are not of interest for classifying. To address this consideration, we propose a loss function that uses the labeled negative examples of the same class as the anchor and thus as additional positive examples. Incorporating additional information as positive examples should proceed carefully. In particular, in a class distribution mismatch scenario, the overall semantic data structure might be easily broken when an OOD instance is selected as the positive example of the corresponding ID anchor and vice versa. To this end, we aggregate the labeled data, which are indeed ID, into positive examples.

The loss function is formulated as follows:
\begin{equation}
\label{eqn:id_loss}
    \begin{split}
    \small
    & \mathcal{L}^{ID}_i =
    \\ & -\tfrac{I(i)}{|P(i;t)|} \log\sum_{p\in P(i;t)} \tfrac
    {\exp({{z_i^a} \cdot {z_p^+}}/\tau)} {\exp({{z_i^a}\cdot{z_i^+}}/\tau) + \sum_{k=1}^{K}\exp({z_i^a}\cdot{z_{i,k}^-}/\tau)},
    \end{split}
\end{equation}

where $I(i)$ is an indicator function that $I(i)=1$ if an anchor $x_i^a \in \mathcal{D}_L$, otherwise $I(i)=0$. The proposed loss function is only applicable when the anchor has class information. $P(i;t)$ denotes an index set of labeled keys of the memory queue at iteration $t$ having the same class as the anchor. As the keys in a memory queue dynamically change along training iteration, $P(i;t)$ also changes accordingly.

$\mathcal{L}^{ID}$ helps the model learn data representations that are more beneficial for downstream classification tasks. On the other hand, $\mathcal{L}^{MoCo}$, which is applied to all data, performs instance discrimination, enabling the learning of more nuanced features within each class. We train the proposed model using a weighted sum of the two losses, balancing the need for detailed intra-class feature learning and effective data representation for classification.

The memory queue is a key feature of MoCo, allowing for the storage of past representations generated by the key network. At the end of every training iteration, the current representations $\{z_i^+\}_{i=1}^B$ are enqueued to the memory queue, and the most outdated and the least consistent $B$ representations in the queue are dequeued, where $B$ refers to the mini-batch size. With this operation, MoCo maintains the keys on-the-fly during training with manageable extra computations. We modified the memory queue to be appropriate for addressing the safe SSL problem. Our modification allows the memory queue to store not only the representations but also their corresponding class information. It equips the memory queue with the capability to pair each stored representation with its class label, offering a more organized and informative approach to managing representations.

In our method, we recognized the potential for overfitting due to the reuse of the same labeled data in both the pre-training and fine-tuning phases. This risk can be particularly pronounced in SSL because of the limited amount of labeled data available. To address this, we applied a coefficient schedule $w(t)$ to gradually reduce the influence of the proposed loss function, $\mathcal{L}^{ID}$, on the model training as the training progresses.

Instance discrimination, which is the focus of $\mathcal{L}^{MoCo}$, represents a more challenging task than class discrimination, which is the focus of $\mathcal{L}^{ID}$. If the influence of $\mathcal{L}^{ID}$ remains strong throughout the training process, it may hinder the model's ability to adequately learn from $\mathcal{L}^{MoCo}$. In the initial stages of training, $\mathcal{L}^{ID}$ plays a critical role in ensuring that class representations are effectively clustered. However, as training progresses, reducing its influence becomes reasonable to prevent the model from being overly tuned to $\mathcal{L}^{ID}$ at the expense of $\mathcal{L}^{MoCo}$.

Therefore, a schedule of the balance coefficient is essential. For simplicity, we adopted a function that starts at one and decreases linearly to zero at epoch $t_{end}$ for $w(t)$, which constantly remains zero after $t_{end}$. The coefficient schedule is designed to adjust the balance between the two types of learning as training progresses, thereby facilitating robust generalization and reducing the risk of overfitting. The final loss function is formulated as follows:
\begin{equation}
\label{eqn:fin_loss}
    \mathcal{L}_i = \mathcal{L}_{i}^{MoCo} + \alpha w(t) \mathcal{L}_{i}^{ID},
\end{equation}
where $\alpha$ is a balancing hyperparameter. After pre-training, the final SSL classifier is trained on $\mathcal{D}_L$ through the fine-tuning phase. Details of the fine-tuning are illustrated in Section \ref{sec:experiments}.

\section{Experiments}
\label{sec:experiments}
\subsection{Data}
\label{sec:experiments_data}

\begin{table*}[!tb]
\renewcommand\arraystretch{1.2}
\centering
\caption{Data configuration of CIFAR-10 for various mismatch ratios.}
\label{table:cifar10}
\resizebox{0.90\textwidth}{!}{
\begin{tabular}{ll|c|cccccc|cccc}
\hline\hline
\multicolumn{1}{c}{\multirow{2}{*}{Split}} & \multicolumn{1}{c|}{\multirow{2}{*}{Type}} & \multirow{2}{*}{\begin{tabular}[c]{@{}c@{}}Mismatch\\ Ratio\end{tabular}} & \multicolumn{6}{c|}{In-Distribution}      & \multicolumn{4}{c}{Out-of-Distribution} \\
\multicolumn{1}{c}{}                       & \multicolumn{1}{c|}{}                      &                                                                           & Bird & Cat  & Deer & Dog  & Frog & Horse & Airplane    & Car    & Ship   & Truck   \\
\hline
Training                                      & Labeled                                   & -                                                                         & 400  & 400  & 400  & 400  & 400  & 400   &             &        &        &         \\
\hline
Training                                      & Unlabeled                                 & 0\%                                                                       &      &      & 4,200 & 4,200 & 4,200 & 4,200  &             &        &        &         \\
                                           &                                           & 25\%                                                                      &      &      &      & 4,200 & 4,200 & 4,200  & 4,200        &        &        &         \\
                                           &                                           & 50\%                                                                      &      &      &      &      & 4,200 & 4,200  & 4,200        & 4,200   &        &         \\
                                           &                                           & 75\%                                                                      &      &      &      &      &      & 4,200  & 4,200        & 4,200   & 4,200   &         \\
                                           &                                           & 100\%                                                                     &      &      &      &      &      &       & 4,200        & 4,200   & 4,200   & 4,200    \\
\hline
Validation                                 & Labeled                                   & -                                                                         & 400  & 400  & 400  & 400  & 400  & 400   &             &        &        &         \\
\hline
Testing                                       & Labeled                                   & -                                                                         & 1,000 & 1,000 & 1,000 & 1,000 & 1,000 & 1,000  &             &        &        &        \\
\hline\hline
\end{tabular}
}
\end{table*}

For a thorough and fair comparison, we followed the experimental settings and evaluation protocol for SSL with class distribution mismatch proposed by \cite{oliver2018realistic} and \cite{chen2020semi}. We used three benckmark datasets for our experiments: CIFAR-10, CIFAR-100, and Tiny ImageNet. Both CIFAR-10 and CIFAR-100 consists of 60,000 color images of size $32\times32$, from which 50,000 and 10,000 images were used for training and testing, respectively. The images were equally divided according to the number of classes, 10 or 100. Tiny ImageNet consists of 200 classes, with 500 traiing images and 50 testing images for each class. Because the size of all Tiny ImageNet color images is $64\times64$, all Tiny ImageNet samples were resized to $32\times32$, the same size as CIFAR-10 and CIFAR-100.

First, we modified CIFAR-10 to examine the performance of the proposed method within various mismatch ratios. We designed a 6-class classification task to classify animals. Among the classes of CIFAR-10, bird, cat, deer, dog, frog, and horse were considered ID classes. On the other hand, airplane, car, ship, and truck were considered OOD classes. For each class, 400 images were used as training labeled data ($\mathcal{D}_L$), 400 images were used as validation data, and the rest were used as training unlabeled data ($\mathcal{D}_U$). We set the mismatch ratio from 0\% to 100\% by changing the classes constituting the unlabeled data, as shown in Table \ref{table:cifar10}. For instance, for a mismatch ratio of 50\%, the classes of the unlabeled data consisted of frog, horse, airplane, and car. Second, we modified CIFAR-100 and Tiny ImageNet to evaluate the performances under class distribution mismatch in larger class spaces. From 100 of CIFAR-100 classes, the 1$\sim$50 classes were used as labeled classes, and 26$\sim$75 classes were used as unlabeled classes. From 200 Tiny ImageNet classes, the 1$\sim$100 classes were used as labeled classes, and the 51$\sim$150 classes were used as unlabeled classes. Therefore, the mismatch ratio of the modified datasets is 50\%. For each class, 100 images per class were used as $\mathcal{D}_L$, 100 images were used as validation data, and the rest were used as $\mathcal{D}_U$.

To more extensively evaluate our method, we created a cross-dataset scenario by combining CIFAR-100 and Tiny ImageNet. Following the data construction method proposed by UASD \cite{chen2020semi}, CIFAR-100 was used as labeled data and Tiny ImageNet was used as unlabeled data. Because Tiny ImageNet contains the same or similar classes as CIFAR-100, we can create a class distribution mismatch scenario. We denoted the dataset as CIFAR-100+Tiny ImageNet and the mismatch ratio is 86.5\%.

For both SSL and SSCL, model performances can change quite a bit depending on which data augmentation strategy is applied to which method. Therefore, it is important to carefully select the appropriate data augmentation. To directly demonstrate the effectiveness of using an SSCL approach for safe SSL, we used similar data augmentation strategies for SSL and the proposed method. For SSL methods, we used a standard strategy consisting of random resizing, random cropping, random horizontal flipping, and adding Gaussian noise \cite{oliver2018realistic}. For the proposed method, random color jittering and grayscale conversion are substituted for the Gaussian noise \cite{chen2020improved}. In SSCL, applying the color-related data augmentations is important. Because the cropped images from an identical image share the nearly same color histogram, the model focuses on merely memorizing the histogram, resulting in poor representations \cite{chen2020simple}. The aforementioned strategies are suitable and are generally used for the methods described in Section \ref{sec:experiments_params}.

\subsection{Model Configuration and Training Parameters}
\label{sec:experiments_params}
We compared the proposed method with a supervised model trained on only labeled data, six representative deep SSL methods, and four state-of-the-art safe SSL methods. The SSL methods are pseudo-label \cite{lee2013pseudo}, VAT \cite{miyato2018virtual}, $\Pi$-Model \cite{sajjadi2016regularization}, temporal ensembling \cite{laine2016temporal}, mean teacher \cite{tarvainen2017mean}, and SWA \cite{athiwaratkun2018there}. Although these methods have achieved great results, they operate under the conventional SSL assumption that the class distribution mismatch does not exist. MTCF \cite{yu2020multi}, DS3L \cite{guo2020safe}, UASD \cite{chen2020semi}, and TOOR \cite{huang2022they} are recently proposed safe SSL methods that consider the challenging scenario in which the identical distribution assumption is violated. MTCF considers that the ID and OOD data are originated from two different domains and distinguishes them with curriculum learning. DS3L uses an extra weight network to control the amount of influence on a training classifier per unlabeled data observation while UASD adopts a self-distillation framework to robustly calculate confidence scores, which are consequently used to dynamically filter out the OOD data during training. TOOR reuses the OOD samples that have similar representations to labeled data by positioning them on similar feature space with adversarial domain adaptation. We conducted experiments to directly demonstrate that learning general data representations based on SSCL can enhance the model performance without unlabeled OOD data detection.

SSL methods follow the unified experimental setting and training procedure proposed by \cite{oliver2018realistic}. Unless otherwise noted, all experimental settings for SSL methods, including hyperparameters and training procedures, are the same as those of \cite{oliver2018realistic}. We used WRN-28-2, i.e., Wide ResNet with depth 28 and width 2 \cite{zagoruyko2016wide} including batch normalization and leaky ReLU \cite{xu2015empirical} with a slope of 0.1 as a base network for SSL and safe SSL methods. The dimension of the last classification layer is revised from ten to $C$, where $C$ is the number of target classes ($C=6, 50, 100$, and $100$ on CIFAR-10, CIFAR-100, Tiny ImageNet, and CIFAR-100+Tiny ImageNet, respectively). Safe SSL methods follow the experimental settings provided by each corresponding study. Each model is trained for 500,000 iterations with a batch size of 100.

We used the default network structure and hyperparameters recommended in MoCo-v2 \cite{chen2020improved} unless otherwise specified. We selected ResNet-50 \cite{he2016deep} as the base architecture of the proposed method. The fully-connected layers were replaced with a multilayer perceptron to build key network $f_{\theta_k}$ and query network $f_{\theta_q}$. The output dimension of the multilayer perceptron is 128 and the outputs are L2-normalized. Because ResNet-50 is originally designed for the ImageNet dataset, which has a bigger image size than CIFAR-10 and CIFAR-100, we slightly modified the architecture. The first $7\times7$ convolution filter with a stride of two and zero padding of three was replaced with a $3\times3$ convolution filter with a stride of one and zero padding of one. We also removed the max-pooling layer, and as a result, the network produces a $4\times4$ feature map. The MoCo originally uses a shuffling batch normalization technique to avoid easily obtaining positive pairs only with batch statistics as a hint. Without this technique, the model might learn degenerated solutions by easily finding positive pairs \cite{he2020momentum}. Because it was originally developed to work only with multiple GPUs, we used ghost normalization \cite{hoffer2017train}. The ghost normalization divides a batch into smaller sub-batches and calculates the batch statistics independently. In our experiments, we split a batch into eight sub-batches because the original MoCo uses eight GPUs.

For all experiment, we set $\tau=0.2$. We used $m=0.950$ and $K=4096$ for CIFAR-10 and CIFAR-100, while $m=0.999$ and $K=8192$ were used for Tiny ImageNet. $\alpha$ and $t_{end}$ are critical hyperparameters for balancing two loss terms. Therefore, we show the experimental results with different choices of $\alpha$ and $t_{end}$ in Section \ref{sec:experiments_results}. With these hyperparemters, we pre-trained the model for 1,000 epochs with a batch size of 256. We used an SGD optimizer with a learning rate of 0.03 and a momentum of 0.9. Started from the initial learning rate, the learning rate gradually decreases to zero by following a half-period cosine schedule. All experiments were conducted on a single NVIDIA TITAN RTX GPU.

\subsection{Evaluation Protocol}
\label{sec:experiments_eval}
After pre-training, we evaluated the quality of representations by two commonly used protocols: simple weighted $k$-nearest neighbor classification ($k$-NN) and linear classification \cite{wu2018unsupervised, caron2021emerging, caron2020unsupervised}. First, we froze the pre-trained model to compute the representation vectors of the training data for $k$-NN classification. Then, the $k$-NN classifier matches the label of a validation or test image to the label voted by $k$-nearest training images. This evaluation can be conducted without any other hyperparameter tuning or additional model training. Therefore, we can continuously monitor the $k$-NN accuracy not only after the pre-training has ended, but also during the pre-training. We used $k$ of 5 and 200. The accuracy obtained with small $k$ values indicates how well semantically similar ID instances are clustered, while the accuracy with large $k$ values indicates how linearly separable each class is.

Second, for the linear classification, we froze the weights of the pre-trained network. The multilayer perceptron was replaced with a single-layer softmax classifier with a dimension of $C$. We trained the classifier on $\mathcal{D}_L$ for 100 epochs. An SGD optimizer with a learning rate of 30 was used, and the learning rate gradually decreases to zero using a half-period cosine schedule. We only applied horizontal flipping and random cropping for training the linear classifier. No data augmentation was applied to the testing dataset.

Finally, we conducted a supervised fine-tuning to evaluate the classification performance of the proposed method. The fine-tuning resembles the approach used in the linear classification, where we substituted the multilayer perception with a single-layer softmax classifier. It is important to note that during the fine-tuning phase, we did not freeze the pre-trained network; instead, we allowed further training and adjustments on weights to occur. This approach was employed to ensure that the model adpats more effectively to the specific characteristics of $\mathcal{D}_L$, resulting in high classification accuracy. We fine-tuned the network on $\mathcal{D}_L$ for 100 epochs. Except for the learning rate of 0.03, optimizer, learning rate schedule, and data augmentations used the same settings as those used for the linear classification. Consequently, we can achieve a trained safe SSL classifier that plays the same role as $h(x)$ in Equation \ref{eqn:ssl}. We reported the average performances with standard deviations in parentheses over five runs for all experiments.

\subsection{Results}
\label{sec:experiments_results}
First of all, we investigated the effect of hyperparameters of $\alpha$ and $t_{end}$ for pre-training. In general, the balancing hyperparamters used for adding a new loss term strongly affects model performance. We conducted experiments on CIFAR-10 with a mismatch ratio of 50\% varying the values of $\alpha$ and $t_{end}$. The scenario was chosen as an intermediate level of difficulty among various mismatch scenarios. Tables \ref{table:alpha_end_5nn}, \ref{table:alpha_end_200nn}, and \ref{table:alpha_end_linear} show the $k$-NN and linear classification accuracy over different values of $\alpha$ and $t_{end}$. $t_{end}=none$ means that no schedule is applied and the value of $w(t)$ is always one during pre-training. Additionally, we report the performances of MoCo that uses only $\mathcal{L}^{MoCo}$ as a baseline.

\begin{table}[!t]
\renewcommand\arraystretch{1.0}
\centering
\caption{5-NN classification accuracy obtained over different values of $\alpha$ and $t_{end}$ on CIFAR-10 with a mismatch ratio of 50\%. \textbf{Boldface} values represent the best performance (highest accuracy) for each configuration.}
\label{table:alpha_end_5nn}
\resizebox{1.00\columnwidth}{!}{
{
\begin{tabular}{c|cccc}
\hline\hline
\multirow{2}{*}{$\alpha$} & \multicolumn{4}{c}{$t_{end}$}                                   \\
                       & \textit{none}         & 100          & 200                   & 300          \\
\hline
3                      & 74.53 (0.61) & 70.12 (0.50) & 74.23 (0.39)          & 74.88 (0.66) \\
2                      & 73.35 (0.64) & 70.45 (0.66) & \textbf{75.02 (0.49)} & 74.98 (0.49) \\
1                      & 71.97 (0.66) & 70.13 (0.55) & 74.43 (0.53)          & 74.62 (0.57) \\
0.5                    & 72.12 (0.54) & 67.70 (0.62) & 73.22 (0.60)          & 73.40 (0.43) \\
0.25                   & 73.55 (0.55) & 67.08 (0.44) & 69.73 (0.51)          & 72.07 (0.41) \\
\hline
MoCo                   & \multicolumn{4}{c}{66.08 (0.63)}                                   \\
\hline\hline
\end{tabular}
}
}
\end{table}

\begin{table}[!t]
\renewcommand\arraystretch{1.0}
\centering
\caption{200-NN classification accuracy obtained over different values of $\alpha$ and $t_{end}$ on CIFAR-10 with a mismatch ratio of 50\%. \textbf{Boldface} values represent the best performance (highest accuracy) for each configuration.}
\label{table:alpha_end_200nn}
\resizebox{1.0\columnwidth}{!}{
{
\begin{tabular}{c|cccc}
\hline\hline
\multirow{2}{*}{$\alpha$} & \multicolumn{4}{c}{$t_{end}$}                                   \\
                       & \textit{none}         & 100          & 200                   & 300          \\
\hline
3                      & 76.45 (0.59) & 72.97 (0.49) & 77.53 (0.37)          & 77.81 (0.63) \\
2                      & 76.15 (0.61) & 73.22 (0.61) & \textbf{78.07 (0.48)} & 77.92 (0.54) \\
1                      & 75.18 (0.67) & 72.88 (0.51) & 77.75 (0.54)          & 77.12 (0.49) \\
0.5                    & 74.57 (0.50) & 70.70 (0.60) & 75.10 (0.61)          & 77.57 (0.37) \\
0.25                   & 75.23 (0.50) & 70.07 (0.42) & 72.17 (0.49)          & 74.43 (0.40) \\
\hline
MoCo                   & \multicolumn{4}{c}{69.34 (0.63)}                                  \\
\hline\hline
\end{tabular}
}
}
\end{table}

\begin{table}[!t]
\renewcommand\arraystretch{1.0}
\centering
\caption{Linear classification accuracy obtained over different values of $\alpha$ and $t_{end}$ on CIFAR-10 with a mismatch ratio of 50\%. \textbf{Boldface} values represent the best performance (highest accuracy) for each configuration.}
\label{table:alpha_end_linear}
\resizebox{1.0\columnwidth}{!}{
{
\begin{tabular}{c|cccc}
\hline\hline
\multirow{2}{*}{$\alpha$} & \multicolumn{4}{c}{$t_{end}$}                                   \\
                       & \textit{none}         & 100          & 200                   & 300          \\
\hline
3                      & 69.79 (0.27) & 75.83 (0.37) & 77.42 (0.51)          & 77.40 (0.45) \\
2                      & 70.36 (0.29) & 75.41 (0.41) & \textbf{78.95 (0.52)} & 77.72 (0.60) \\
1                      & 63.74 (0.31) & 75.57 (0.53) & 77.58 (0.41)          & 77.31 (0.37) \\
0.5                    & 57.24 (0.30) & 74.16 (0.30) & 76.88 (0.39)          & 77.54 (0.30) \\
0.25                   & 58.39 (0.33) & 74.57 (0.54) & 75.16 (0.43)          & 75.02 (0.51) \\
\hline
MoCo                   & \multicolumn{4}{c}{73.74 (0.37)}                                   \\
\hline\hline
\end{tabular}
}
}
\end{table}

\begin{table}[!t]
\renewcommand\arraystretch{1.2}
\centering
\caption{Linear classification accuracy obtained over different dataset scenarios. Best numbers for each scenario is \textbf{boldfaced}.}
\label{table:linear}
\resizebox{1.0\columnwidth}{!}{
\begin{tabular}{c|ccccc|c|c}
\hline\hline
\multirow{2}{*}{Model}    & \multicolumn{5}{c|}{CIFAR-10}                                             & CIFAR-100       & Tiny ImageNet    \\
                          & 0\%             & 25\%            & 50\%            & 75\%            & 100\%           & 50\%            & 50\%            \\
\hline
\multirow{2}{*}{MoCo}     & 77.78           & 75.96           & 73.82           & 72.17           & 70.47           & 58.92           & 40.97           \\
                          & (0.18)          & (0.61)          & (0.36)          & (0.27)          & (0.54)          & (0.62)          & (0.37)          \\
\multirow{2}{*}{Proposed} & \textbf{82.03}  & \textbf{80.54}  & \textbf{78.91}  & \textbf{78.03}  & \textbf{76.29}  & \textbf{63.73}  & \textbf{45.98}  \\
                          & \textbf{(0.25)} & \textbf{(0.30)} & \textbf{(0.62)} & \textbf{(0.71)} & \textbf{(0.67)} & \textbf{(0.60)} & \textbf{(0.64)} \\
\hline\hline
\end{tabular}
}
\end{table}

Tables \ref{table:alpha_end_5nn}, \ref{table:alpha_end_200nn}, and \ref{table:alpha_end_linear} present the representation qualities evaluated through 5-NN, 200-NN, and linear classification, respectively. Notably, our method consistently outperforms the baseline MoCo across various settings of $\alpha$ and $t_{end}$. The NN-based classification accuracy, assessed during the model's pre-training phase, demonstrates that our approach that utilizes additional class information can more effectively bring representations of the same class closer together compared to MoCo, which solely relies on instance discrimination. The proposed model achieved its highest performance with $\alpha=2$ and $t_{end}=200$, marking a significant improvement of approximately 9\% over MoCo. This highlights the effectiveness of our coefficient schedule, enabling the pre-training phase to yield representations that are more beneficial to downstream classification tasks. Additionally, Table \ref{table:alpha_end_linear} shows that without the coefficient schedule, i.e., $t_{end}=none$, there is a decline in the linear evaluation performance of the proposed model, contrasting with the NN-based results. This performance reduction is attributed to overfitting, which occurs due to the use of a limited amount of labeled data in both the pre-training and linear classification phases. This outcome emphasizes the necessity of the coefficient schedule for maintaining the model's generalizability, as it prevents over-adaption to the small labeled data. Based on these results, we fixed $\alpha$ to 2 and $t_{end}$ to 200 in all remaining experiments.

With the fixed hyperparameters $\alpha=2$ and $t_{end}=200$, we compared the performance of MoCo and the proposed method in various datasets. We evaluated the yielded representation quality by linear classification; the results are shown in Table \ref{table:linear}. The results confirmed that the proposed method yielded better representations than MoCo in all cases, showing performance improvements of about 5\% on average. In the experiments that changed the mismatch ratio in CIFAR-10, the model's overall performance is degraded because the larger the mismatch ratio, the more difficult the scenario is. Recall that there is no ID data in the unlabeled data when the mismatch ratio is 0\%. Nevertheless, the proposed method performed well with substantial performance gaps. The proposed method also yielded better results in CIFAR-100 and Tiny ImageNet.

\begin{figure*}[htp]
\centering
    \subfloat[$\mathcal{L}^{MoCo}$]{\includegraphics[width=0.25\linewidth, keepaspectratio]{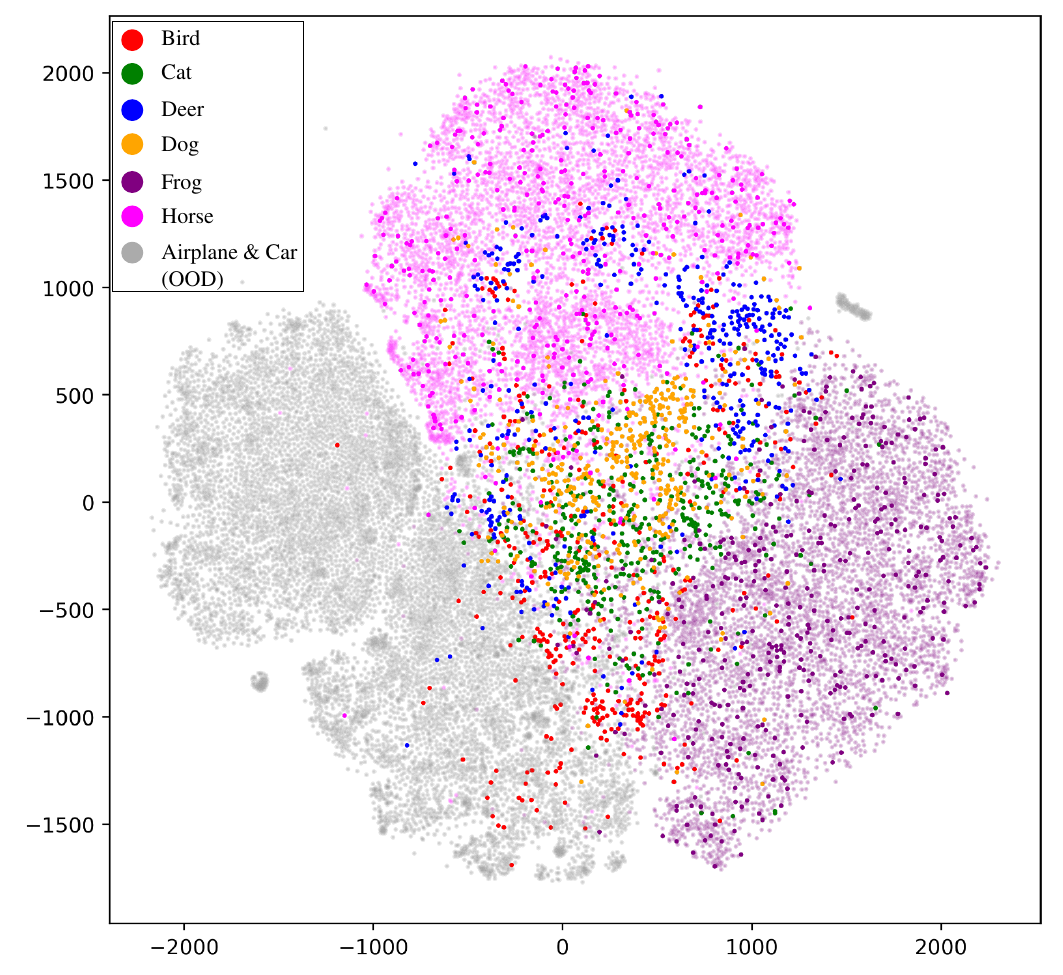}\label{fig:moco-last}}
    \subfloat[$\mathcal{L}^{MoCo}+\alpha\mathcal{L}^{ID}$]{\includegraphics[width=0.25\linewidth, keepaspectratio]{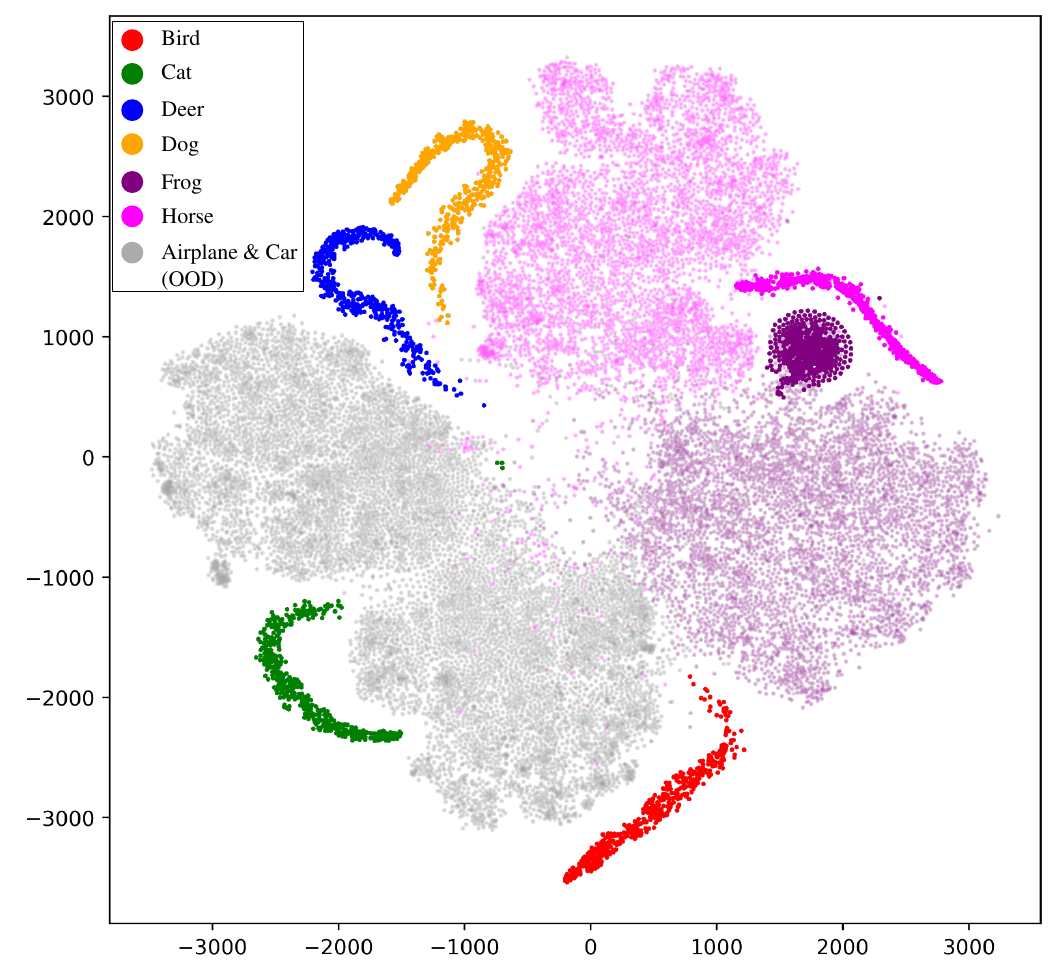}\label{fig:id-last}}
    \subfloat[$\mathcal{L}^{MoCo}+\alpha w(t) \mathcal{L}^{ID}$]{\includegraphics[width=0.25\linewidth, keepaspectratio]{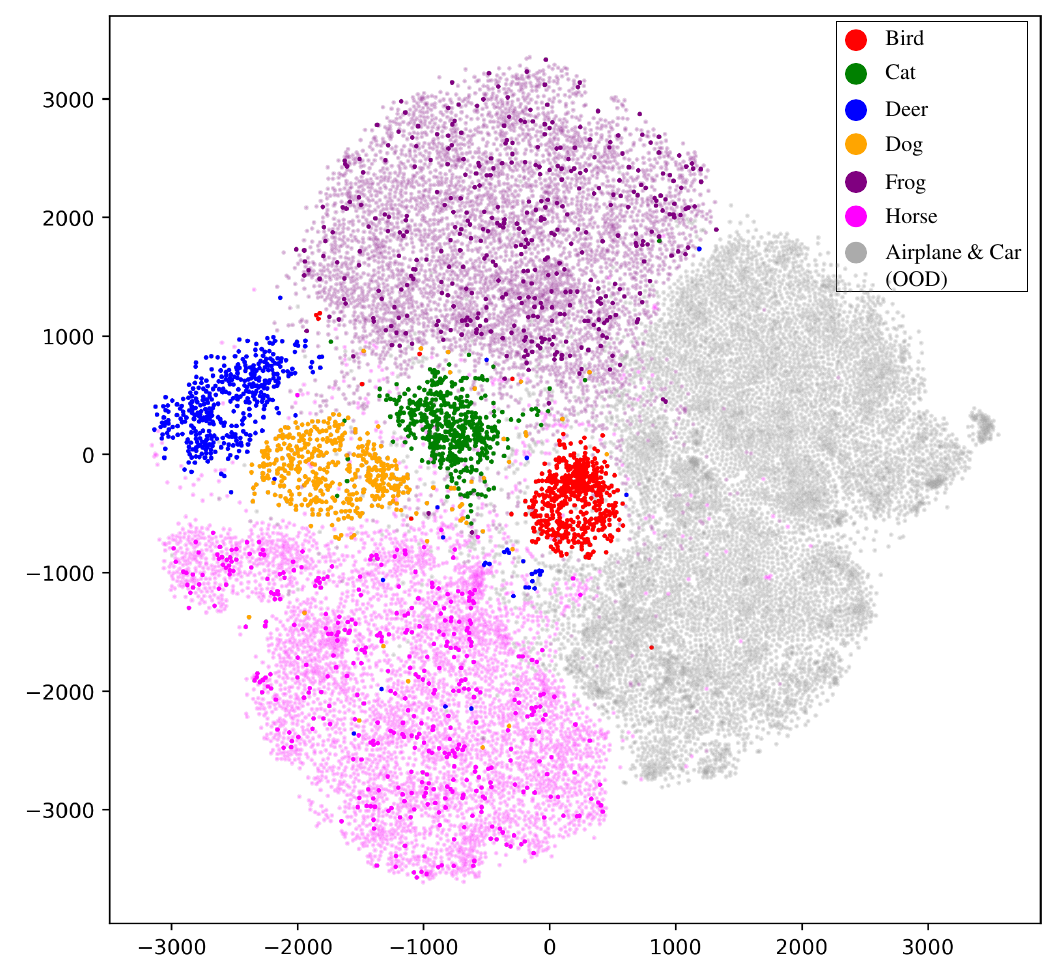}\label{fig:proposed-last}}
    \caption{t-SNE visualization of learned representations. (a) Vanilla MoCo; (b) The proposed loss function without schedule added; (c) The proposed method with schedule added. Colored, lightly colored, and grey points refer to labeled ID, unlabeled ID, and unlabeled OOD data, respectively.}
    \label{fig:tsne-compare}
\end{figure*}

To further assess the representation quality, we visualized the data representations to analyze how the proposed loss function and coefficient schedule affected model training. The representations were obtained from CIFAR-10 with a mismatch ratio of 50\%. A vanilla MoCo, a model in which the proposed loss function is added to MoCo without a schedule, and the proposed model using a schedule were compared. The 128-dimensional representations obtained from each model were reduced to two-dimension vectors by t-SNE for visualization \cite{van2008visualizing}. Figure \ref{fig:tsne-compare} shows the t-SNE results of each model. Note that colored, lightly colored, and grey points refer to labeled ID, unlabeled ID, and unlabeled OOD data, respectively. As shown in Figure \ref{fig:moco-last}, when vanilla MoCo was used, the labeled ID data of different classes overlapped. Moreover, bird, cat, deer, and dog, which are ID classes without unlabeled data, were not represented well. In Figure \ref{fig:id-last}, the labeled ID data is well-grouped by class. However, the labeled data of frog and horse is excessively clustered, so that the unlabeled data of the same class is located in different feature spaces. If a classification boundary is created in this feature space, overfitting occurs. It is also aligned with the results in Table \ref{table:alpha_end_linear}. Figure \ref{fig:proposed-last} shows that the proposed method produces a suitable feature space for the downstream classification task. The ID data is not only well-clustered but is also well-scattered while reflecting the smooth manifold structure. An analysis of Figures \ref{fig:id-last} and \ref{fig:proposed-last} reveals notable differences. Without the proposed coefficient schedule, the model tends to focus more on clustering data, rather than developing sophisticated representations through individual instance discrimination, leading to overfitting.

\begin{figure*}[htp]
\centering
    \subfloat[epoch=1]{\includegraphics[width=0.25\linewidth, keepaspectratio]{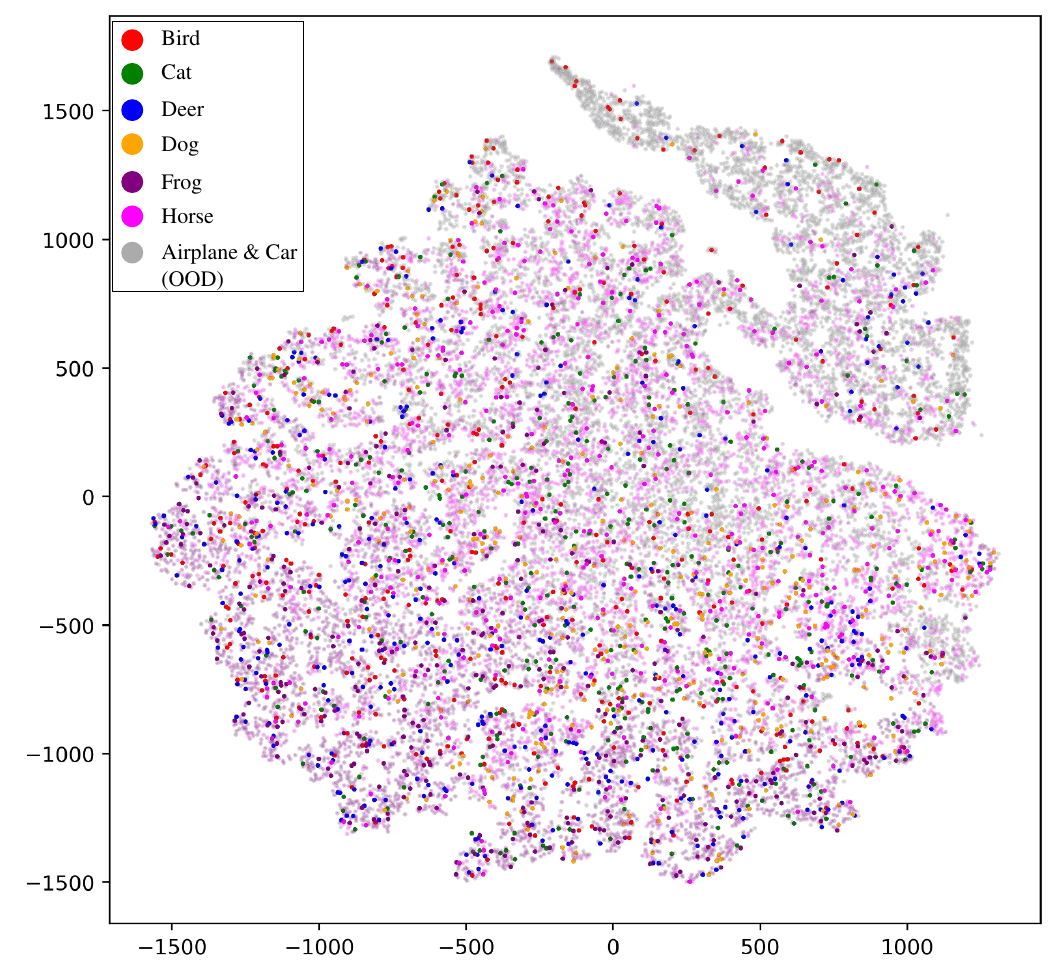}\label{fig:epoch-1}}
    \subfloat[epoch=100]{\includegraphics[width=0.25\linewidth, keepaspectratio]{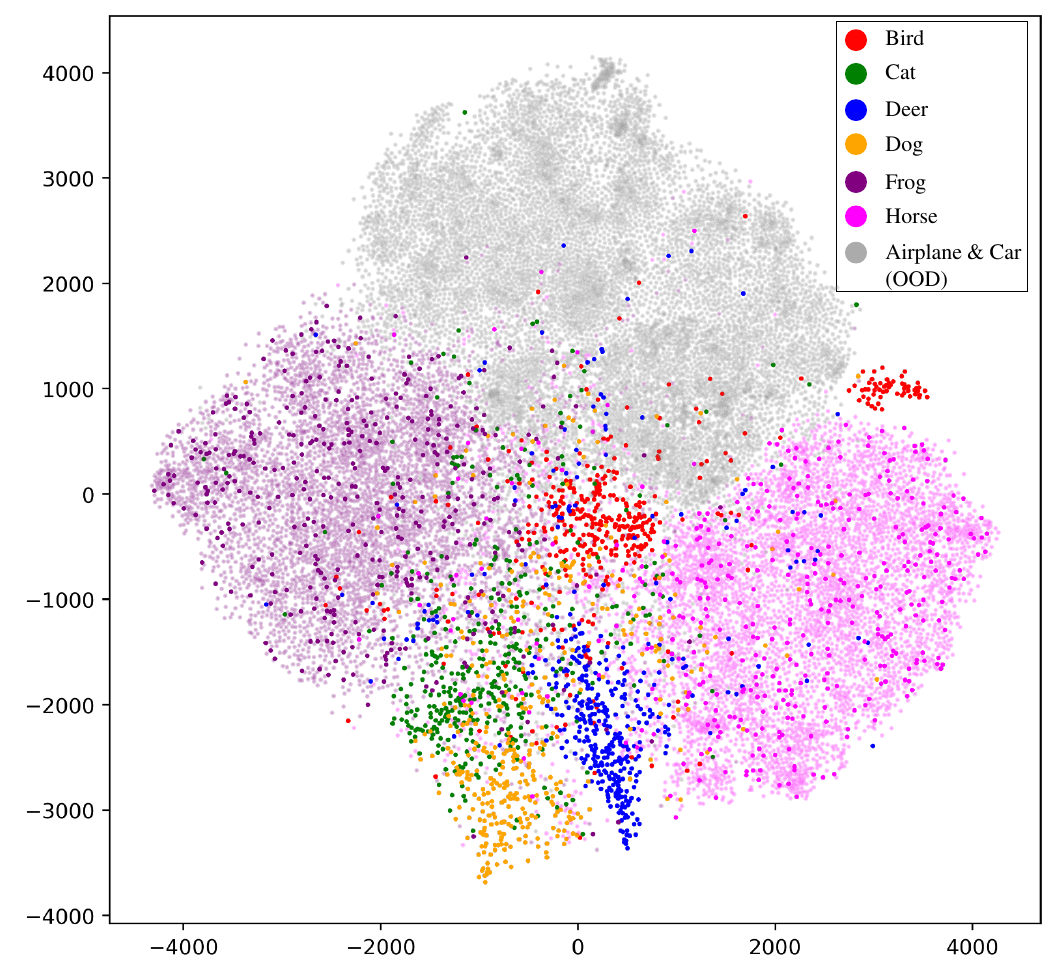}\label{fig:epoch-100}}
    \subfloat[epoch=200]{\includegraphics[width=0.25\linewidth, keepaspectratio]{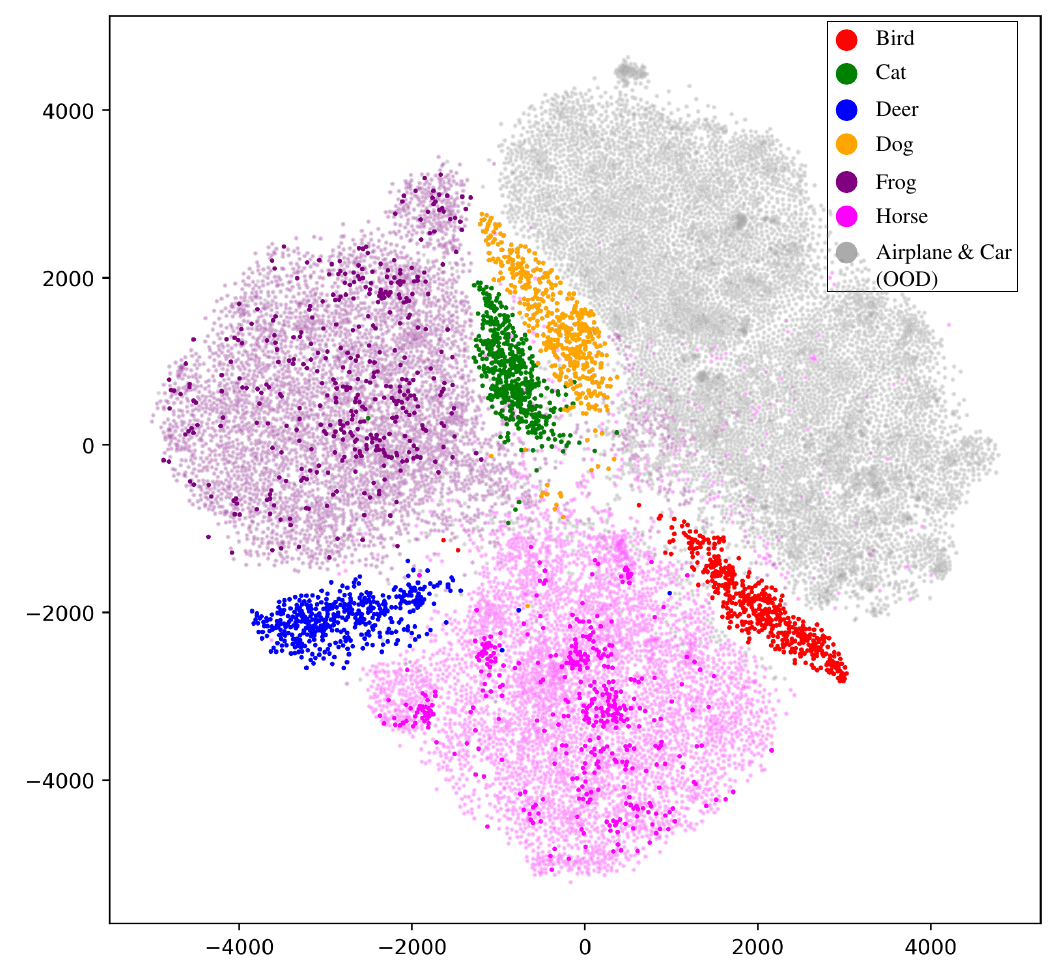}\label{fig:epoch-200}}
    \subfloat[epoch=1,000]{\includegraphics[width=0.25\linewidth, keepaspectratio]{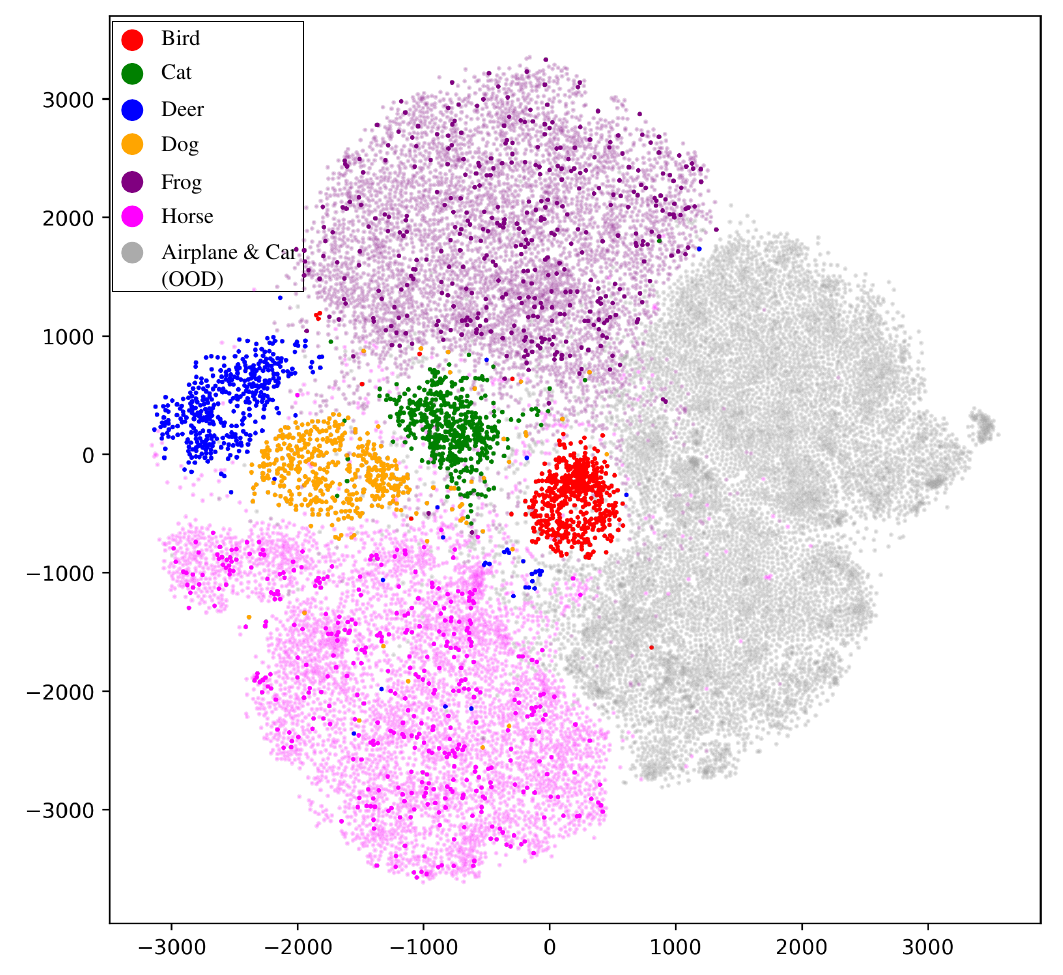}\label{fig:epoch-1000}}
    
    \caption{t-SNE visualization of data representations as the proposed model is trained. $\alpha$ is set to 2 and $t_{end}$ is set to 200.}
    \label{fig:tsne-evolving}
\end{figure*}

Figure \ref{fig:tsne-evolving} shows the evolution of the feature space as our model undergoes training for 1000 epochs. A key observation is the increasingly distinct information of clusters for each class, particularly evident until the completion of the coefficient schedule, as shown in Figure \ref{fig:epoch-200}. Following this period, where $\mathcal{L}^{ID}$ reduces to zero and no longer influences the training, the model starts to spread out the representations more evenly, learning the fundamental semantic data structure. This transition highlights the model's capability to uncover hidden data patterns. Initially, by emphasizing $\mathcal{L}^{ID}$ during the early stage of pre-training, distinct clusters among labeled positive examples are formed. Once $\mathcal{L}^{ID}$ is phased out, the focus shifts towards recognizing the finer details of each individual instance.

Finally, Table \ref{table:classification} presents the classification results. MoCo and the proposed method were fine-tuned as demonstrated in Section \ref{sec:experiments_eval}. Recall that \textit{Supervised} refers to a model trained on only labeled data. Pseudo-label, VAT, $\Pi$-Model, temporal ensembling, mean teacher, and SWA are SSL methods. MTCF, DS3L, UASD, and TOOR are safe SSL methods. MoCo and the proposed method are SSCL methods. Overall, classification accuracy is good in the order of SSCL, safe SSL, SSL, and \textit{Supervised}. However, as the mismatch ratio is increased, the SSL methods sometimes performed worse than \textit{Supervised}. In particular, this phenomenon was greatly pronounced in the rest of the SSL methods except for SWA. In cases of safe SSL methods, TOOR showed better performances than others in overall.

An interesting and meaningful result is that simply applying MoCo greatly enhanced the classification accuracy in all cases. Compared with UASD, which is the best model among existing studies, MoCo showed a performance improvement of about 1$\sim$2\% in CIFAR-10. The best accuracy of the previous methods when the mismatch ratio is 50\% is 78.67\% with TOOR. This accuracy is similar to the MoCo accuracy of 78.71\% when the mismatch ratio is 100\%. It can be seen that MoCo shows better performance than the existing methods even if it is a much more challenging scenario. MoCo also performed well in CIFAR-100, Tiny ImageNet, and CIFAR-100+Tiny ImageNet, which have the mismatch problem in larger class spaces. In particular, MoCo could achieve a considerable performance improvement in Tiny ImageNet compared to the best previous methods. MoCo, based on the concept of instance discrimination, could yield more performance improvements in Tiny ImageNet than in CIFAR-10 and CIFAR-100 because Tiny ImageNet is a fine-grained dataset in which the characteristics of each class and observation are well segmented. We can confirm that the representations learned on a dataset without excluding OOD data can serve as appropriate initial weights for improving the performance of downstream classification models. Our methods also obtained improvements under cross-dataset scenario, CIFAR-100+Tiny ImageNet.

\begin{table*}[!tb]
\renewcommand\arraystretch{1.6}
\centering
\caption{Fine-tuning classification accuracy obtained over different dataset scenarios. \textbf{Boldface} values represent the best performance (highest accuracy) for each scenario. Standard deviations are shown in parentheses.}
\label{table:classification}
\resizebox{2.0\columnwidth}{!}{
\begin{tabular}{c|ccccc|c|c|c}
\hline\hline
Model & \multicolumn{5}{c|}{CIFAR-10}                                                                                          & CIFAR-100             & TinyImageNet          & CIFAR-100 + Tiny ImageNet  \\
Mismatch Ratio      & 0\%                   & 25\%                  & 50\%                  & 75\%                  & 100\%                 & 50\%                  & 50\%                  & 86.5\%                  \\
\hline
Supervised             & \multicolumn{5}{c|}{76.05 (0.68)}                                                                                      & 59.93 (0.97)          & 38.71 (0.54)          & 51.30 (0.56)          \\
Pseudo-Label           & 78.53 (0.36)          & 75.94 (0.73)          & 74.83 (0.56)          & 74.19 (1.01)          & 73.31 (0.45)          & 56.64 (0.50)          & 37.97 (0.61)          & 46.91 (0.70)          \\
VAT                    & 79.57 (1.17)          & 77.02 (0.49)          & 76.12 (0.66)          & 74.71 (0.74)          & 73.86 (0.82)          & 56.01 (1.08)          & 35.81 (0.87)          & 49.77 (0.59)          \\
Pi-Model               & 78.74 (0.79)          & 75.90 (0.72)          & 74.48 (0.51)          & 74.07 (0.51)          & 73.27 (0.81)          & 56.84 (0.42)          & 38.13 (0.54)          & 47.23 (1.50)          \\
Temporal Ensembling    & 78.27 (0.53)          & 75.81 (0.76)          & 74.32 (0.90)          & 73.51 (0.66)          & 72.70 (0.79)          & 58.43 (0.72)          & 40.21 (0.40)          & 52.30 (0.61)          \\
Mean-Teacher           & 78.28 (0.76)          & 75.81 (1.01)          & 75.05 (1.10)          & 74.10 (1.05)          & 73.11 (0.41)          & 59.22 (1.12)          & 40.57 (0.33)          & 51.64 (1.35)          \\
SWA                    & 78.45 (0.41)          & 76.75 (0.88)          & 76.11 (0.71)          & 75.90 (0.82)          & 75.69 (0.75)          & 62.14 (0.66)          & 42.70 (0.42)          & 54.83 (0.61)          \\
MTCF                   & 78.83 (0.71)          & 77.17 (0.76)          & 76.64 (0.69)          & 76.01 (0.72)          & 74.67 (0.73)          & 62.77 (0.62)          & 42.75 (0.54)          & 56.64 (0.59)          \\
DS3L                   & 79.51 (0.35)          & 78.73 (0.51)          & 77.83 (0.45)          & 76.55 (0.63)          & 76.27 (0.57)          & 63.71 (0.64)          & 42.92 (0.61)          & 55.70 (0.43)          \\
UASD                   & 79.41 (0.51)          & 78.66 (0.52)          & 78.12 (0.69)          & 77.61 (0.48)          & 77.51 (0.90)          & 64.07 (0.60)          & 43.31 (0.78)          & 57.21 (0.32)          \\
TOOR                   & 80.23 (0.46)          & 79.17 (0.49)          & 78.67 (0.34)          & 77.49 (0.44)          & 76.13 (0.62)          & 64.47 (0.53)          & 44.14 (0.70)          & 58.11 (0.37)          \\
\hline
MoCo                   & 82.27 (0.62)          & 81.16 (0.50)          & 80.10 (0.29)          & 78.80 (0.46)          & 78.71 (0.55)          & 66.94 (0.35)          & 50.75 (0.37)          & 59.51 (0.40)          \\
Proposed               & \textbf{82.58 (0.29)} & \textbf{81.64 (0.30)} & \textbf{80.64 (0.23)} & \textbf{80.21 (0.39)} & \textbf{79.82 (0.42)} & \textbf{69.68 (0.20)} & \textbf{53.41 (0.48)} & \textbf{61.39 (0.31)} \\
\hline\hline
\end{tabular}
}
\end{table*}

\begin{table*}[!tb]
\renewcommand\arraystretch{1.6}
\centering
\caption{Classification accuracy of the models trained with strong augmentation. \textbf{Boldface} values represent the best performance (highest accuracy) for each scenario. Standard deviations are shown in parentheses.}
\label{table:classification-strong}
\resizebox{2.0\columnwidth}{!}{
\begin{tabular}{c|ccccc|c|c|c}
\hline\hline
Model           & \multicolumn{5}{c|}{CIFAR-10}          & CIFAR-100 & TinyImageNet & CIFAR-100+TinyImageNet \\
Mismatch Ratio  & 0\%    & 25\%   & 50\%   & 75\%   & 100\%  & 50\%       & 50\%          & 86.5\%                  \\
\hline
FixMatch        & 87.82 (0.20) & 85.23 (0.32) & 84.18 (0.29) & 82.77 (0.41) & 80.98 (0.36) & 72.13 (0.33) & 53.48 (0.33) & 60.71 (0.41)           \\
MoCo-Strong     & 87.77 (0.23) & 85.26 (0.28) & 84.30 (0.24) & 82.84 (0.35) & 80.92 (0.41) & 72.62 (0.40) & 53.66 (0.44) & 60.86 (0.38)           \\
Proposed-Strong & \textbf{87.96 (0.18)} & \textbf{85.89 (0.24)} & \textbf{84.94 (0.28)} & \textbf{83.72 (0.33)} & \textbf{81.79 (0.35)} & \textbf{73.48 (0.32)} & \textbf{55.04 (0.36)} & \textbf{62.33 (0.35)}       \\
\hline\hline
\end{tabular}
}
\end{table*}

The proposed safe semi-supervised contrastive learning method further improved the performance of MoCo. Although the effects of the proposed loss function tended to diminish compared to the linear classification as the weights of a pre-trained network are not frozen for additional training, the proposed method yielded meaningful improvement compared to MoCo. Of note, the performance improvement of the proposed method was more pronounced under challenging scenarios: CIFAR-10 75\% and 100\% with a high mismatch ratio, CIFAR-100 and Tiny ImageNet with a large class space, and CIFAR-100+Tiny ImageNet under cross-dataset scenario. This demonstrates that relatively more performance gains can be achieved when the proposed method using ID data as additional positive examples is applied in challenging scenarios. It is also worth noting that we achieved great results under the restricted experimental settings, including a handful of data augmentations.

In the realm of image classification, it is well-known that strong augmentations like RandAugment \cite{cubuk2020randaugment} or CTAugment \cite{berthelot2019remixmatch} can significantly enhance performance. FixMatch, one of the most representative methods, demonstrates improved results by employing both strong and weak augmentations. However, naively applying strong augmentation during the pre-training phase of SSCL can lead to training failure. This occurs as strong augmentations alter the structural integrity of the images, thereby complicating the retrieval process in SSCL \cite{wang2022contrastive}.

To explore whether strong augmentation could enhance the performance of our model, we conducted a proof-of-concept experiment. We began pre-training with weak augmentation to establish initial network weights, followed by fine-tuning using the FixMatch framework. This approach allows us to determine if representations obtained under class distribution mismatch scenarios could provide a more effective initial solution for fine-tuning. For the baseline, FixMatch was implemented using the same training hyperparameters mentioned in the original paper. Both MoCo and our method were adapted by replacing the backbone architecture with Wide ResNet-28-2 \cite{zagoruyko2016wide}. We used RandAugment for the strong augmentation strategy.

Table \ref{table:classification-strong} displays the classification results for various datasets and mismatch scenarios when strong augmentation is applied. Aligning with the results in Table \ref{table:classification}, our model, denoted as Proposed-Strong, exhibited the best performances compared to competing methods, with particularly notable gains in scenarios featuring large mismatch ratios. MoCo-Strong showed overall performance comparable to FixMatch. However, it is worth noting that MoCo-Strong achieved results similar to FixMatch with only a short period of fine-tuning epochs. Proposed-Strong attained enhanced performance by acquiring representations more focused on the target ID classes during the pre-training phase.

\section{Conclusions}
\label{sec:conclusions}
In this study, we proposed a safe semi-supervised contrastive learning method built upon MoCo. It is the first study to apply a SSCL approach to SSL with OOD data. The proposed method contributes to considerable gains in classification performance in various SSL scenarios involving class distribution mismatches. Because the MoCo learns data representations in terms of instance discrimination, it has the advantage that there is no need to consider the mismatch problem. A new contrastive loss function that utilizes the labeled examples of the same class for additional positive examples was introduced. It helps ID data to be well clustered compared to simple MoCo. We also proposed the use of a coefficient schedule to prevent the model from being overfitted to the small amount of labeled data. To verify the proposed method, we conducted experiments on image classification datasets, CIFAR-10, CIFAR-100, Tiny ImageNet, and CIFAR-100+Tiny ImageNet. We confirmed that applying the proposed loss function with its coefficient schedule greatly enhanced the model performance in terms of both representation quality and classification accuracy.

Although the proposed method showed favorable results, it can be extended in several interesting directions. First, an adaptive scheduling strategy is required. For simplicity, we used a linearly decaying schedule. However, this is also a limitation of our method. The learning curve of the model differs depending on the type of data, and it needs a lot of time and effort to find the appropriate hyperparameter for each dataset. In some cases, labeled information may be needed again in the middle of training. It is thus necessary to design a metric that can quantify the characteristics and develop a method for adaptively adjusting the coefficient of the proposed loss function. Second, we could consider selecting candidates for positive examples from the unlabeled negative examples. The proposed method considers only labeled data as additional positive examples in a safe manner. However, it is evident that a better representation can be obtained if the unlabeled negative examples can also be utilized. Especially in scenarios where label information is extremely rare, it is more necessary to search for useful information in the unlabeled data. Third, it would be interesting to develop a model adept at OOD detection while improving classification performance. From a data collection point of view, there is no need to continuously collect and store unnecessary data. Although the proposed method aimed to achieve high classification performance using the given training data, it will also be necessary to develop an OOD detection model to reduce the data collection cost. Lastly, we can apply the recent algorithms using strong data augmentations in the pre-training phase. To see the effect that can be obtained when applying SSCL, we restricted the study to using a few basic strategies. However, we expect a further performance gain can be achieved by searching for the proper pair of algorithm and data augmentations.

\section*{Acknowledgments}
This research was supported by Brain Korea 21 FOUR, the Ministry of Science and ICT (MSIT) in Korea under the ITRC support program supervised by the IITP (IITP-2020-0-01749), and the National Research Foundation of Korea grant funded by the Korea government (RS-2022-00144190).

\section{References}
\bibliographystyle{IEEEtran}
\bibliography{IEEEabrv,reference}

\end{document}